\documentclass[10pt,twocolumn,letterpaper]{article}

\usepackage{wacv}
\usepackage{times}
\usepackage{epsfig}
\usepackage{graphicx}
\usepackage{amsmath}
\usepackage{amssymb}

\usepackage{xurl}
\usepackage{amsfonts}
\usepackage[utf8]{inputenc}
\usepackage[english]{babel}
\usepackage{amsthm}
\newtheorem{theorem}{Theorem}[section]
\newtheorem{lemma}[theorem]{Lemma}
\theoremstyle{definition}
\newtheorem{definition}{Definition}[section]

\usepackage{mathtools}

\DeclarePairedDelimiter\floor{\lfloor}{\rfloor}
\usepackage{bbm}
\usepackage{diagbox}
\usepackage{multicol}
\usepackage{multirow}
\usepackage{subcaption}
\usepackage{makecell}
\usepackage[
  separate-uncertainty = true,
  multi-part-units = repeat
]{siunitx}
\usepackage{graphicx} 
\usepackage{booktabs}
\def\wacvPaperID{0474} 

\wacvfinalcopy 

\ifwacvfinal
\fi

\ifwacvfinal
\usepackage[breaklinks=true,bookmarks=false]{hyperref}
\else
\usepackage[pagebackref=true,breaklinks=true,colorlinks,bookmarks=false]{hyperref}
\fi

\begin{document}

\title{\vspace{-2.5em}\textit{S2FGAN}: Semantically Aware Interactive Sketch-to-Face Translation}

\author{
Yan Yang$^{1,2}$ \quad Md Zakir Hossain$^{1,2}$ \quad Tom Gedeon$^3$ \quad Shafin Rahman$^4$\\
$^1$ BDSI, Australian National University, Australia \quad $^2$A\&F, CSIRO, Australia \\ $^3$ EECMS, Curtin University,  Australia  \quad $^4$ ECE, North South University, Bangladesh\\
{\tt\small \{Yan.Yang, zakir.hossain\}@anu.edu.au \quad Tom.Gedeon@curtin.edu.au \quad shafin.rahman@northsouth.edu}

}

\maketitle
\thispagestyle{empty}

\begin{abstract}
\vspace{-1em} 
Interactive facial image manipulation attempts to edit single and multiple face attributes using a photo-realistic face and/or semantic mask as input. In the absence of the photo-realistic image (only sketch/mask available), previous methods only retrieve the original face but ignore the potential of aiding model controllability and diversity in the translation process. This paper proposes a sketch-to-image generation framework called \textbf{\textit{S2FGAN}}, aiming to improve users' ability to interpret and flexibility of face attribute editing from a simple sketch. First, to restore a vivid face from a sketch, we propose semantic level perceptual loss to increase the translation quality. Second, we dedicate the theoretic analysis of attribute editing and build attribute mapping networks with latent semantic loss to modify latent space semantics of Generative Adversarial Networks (GANs). 
The users can \textit{command} the model to retouch the generated images by involving the semantic information in the generation process. In this way, our method can manipulate single or multiple face attributes by only specifying attributes to be changed. Extensive experimental results on the CelebAMask-HQ dataset empirically show our superior performance and effectiveness on this task. Our method successfully outperforms state-of-the-art sketch-to-image generation and attribute manipulation methods by exploiting greater control of attribute intensity.
\end{abstract}

\begin{figure}
  \vspace{-1em}
  \includegraphics[width = 0.95\linewidth]{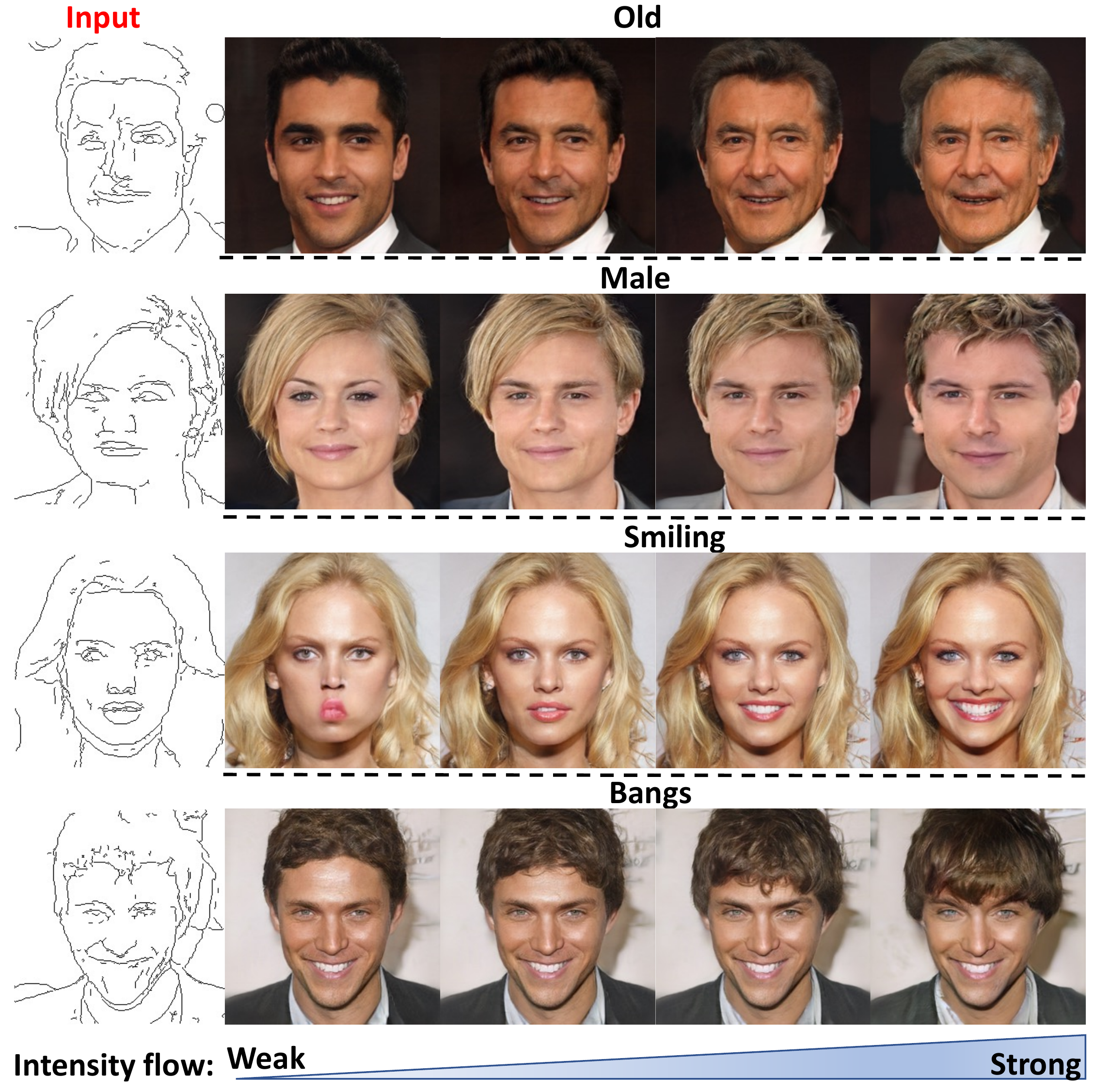}
  \vspace{-1em}
  \caption{\small Our proposed \textbf{\textit{S2FGAN}} framework generates high resolution and photo-realistic face from a face sketch image as {\color{red}{input}} by using semantic conceptual loss and manipulates single or multiple attributes (age, gender, nose, smiles, lips, bangs, and so on) of faces with greater control by using latent semantic loss.
}
\label{fig:coverpage}
\end{figure}

\vspace{-1em}
\section{Introduction}
Generative Adversarial Networks (GANs) \cite{GAN} is one of the emerging techniques for image synthesis and achieves tremendous success in image-to-image (I2I) translation. Now, generating photo-realistic faces from sketches becomes possible by learning pixel-wise correspondence using conditional GANs \cite{cGAN,ContextGAN,DPS,pixel2pixelHD,P2P}. Although such techniques remain successful in enabling a novice artist to restore a face from a sketch, they often fail to control specific facial attributes' intensity. For example, 
a user may wish to add a specific attribute, such as a smile, after generating a face from a drawing and/or s/he may wish to control intensities of facial attributes. For non-artists, modifying the sketches is very hard for describing facial features such as happiness, age, or chubby. However, it is easy to specify those abstract appearances (attribute/semantic) as text input. This paper investigates how a GAN model can increase such users' interpretability and flexibility of attributes editing from a simple sketch.

We identify several limitations in this line of investigation.
\textbf{\textit{(a)}} Existing methods perform attribute editing when the input is a photo-realistic face \cite{StarGAN, AttGAN, STGAN, FaderNetwork,InterFaceGAN}. These methods cannot work in the absence of a photo-realistic image as input.
\textbf{\textit{(b)}} Although methods can successfully add or remove attributes (like age, gender, beard) from photo-realistic faces, they are not able to provide adequate measures to control intensities of attributes. Recent approaches have somewhat addressed this issue, but they mostly fail to preserve subjects identify at higher intensity boundary of attributes. 
Moreover, many methods ignore state-like attributes, such as \textit{chubby}, which are expected to have diverse intensities. \cite{StarGAN,STGAN,AttGAN,FaderNetwork,InterFaceGAN}
\textbf{\textit{(c)}} For attribute editing, users need to specify both the attributes intended to be preserved and changed, increasing the necessity of manual annotations \cite{AttGAN, FaderNetwork, StarGAN}. Recent approaches, InterFaceGAN \cite{InterFaceGAN} and STGAN \cite{STGAN} allow users to edit face attributes by specifying attributes to be changed only. However, none of those methods consider sketch inputs.

This paper presents a novel sketch-to-face GAN framework, \textit{S2FGAN}, to aid the controllability of image attributes in the sketch-to-image generation process. We encapsulate two tasks in a single framework: sketch-to-face generation and face attribute editing.
\textit{For the sketch-to-face generation}, approaches widely use encoder-decoder GAN structures \cite{pixel2pixelHD,deepfacepencil,DeepFaceDrawing}. Since a sketch describes a face's layout but fails to indicate the low-level facial features, the latent code of GAN (semantically encoded low dimensional vectors after discarding redundant information \cite{InterFaceGAN}) struggles to describe low-level facial features, which is essential to restore a photo-realistic face. This problem gets intensified with the increase of output image resolution. To address this problem, we propose a semantic level perceptual loss. It encourages to compose latent code of sketch and ground-truth face indistinguishable. In this way, similar to the latent code of ground-truth face, the resultant latent code of sketch also describes low-level facial features in addition to face layout.
\textit{For attribute editing on the generated face}, we perform conditional manipulation of desired facial attributes without affecting the rest attributes of interests. We present two attribute mapping networks with latent semantic loss to modify semantics in the latent space. Moreover, we provide the theoretical underpinning to establish that proposed attribute mapping networks focus on preserving the semantic and intensity of non-edited attributes. It also helps the GAN decoder correctly constructing the rare attribute combinations (e.g., female and beard). In this way, we preserve identity and edit multiple attributes simultaneously based on only specifying attributes to be changed and manipulate the attribute intensity with greater control. Figure \ref{fig:coverpage} shows sample outputs of our method. 
Comparing with the state-of-the-art methods in Figure \ref{fig:intensitycontrol}, our model provides smooth attributes intensity control. Furthermore, we achieve superior attribute editing and diversity control performance, especially when manipulating multiple attributes, as shown in Figure \ref{fig:multiattributecompare}. Overall, our contributions are summarised below: 
\begin{itemize}\setlength{\itemsep}{-0.3em}

 \item We propose the \textit{S2FGAN} framework for sketch-to-image translation with the facility of face reconstruction, attribute editing, and interactive manipulation of attribute intensity. Our model can work on both single and multiple attribute editing and manipulation problems. Further, users can control intensities of attributes by only specifying/changing target attributes (semantics) and preserving face identity.

 \item We present a semantic level perceptual loss to increase the sketch-to-image translation quality. Our attribute editing models use latent semantic loss, which calibrates face attributes with broader control, diversity, and smoothness.
 
 \item We compare with state-of-the-art (DeepFaceDrawing \cite{DeepFaceDrawing}, DeepFacePencil \cite{deepfacepencil}, Deep Plastic Surgery \cite{DPS}, Pix2PixHD \cite{pixel2pixelHD}) sketch-to-image translation model and attribute editing model (AttGAN \cite{AttGAN}, STGAN \cite{STGAN} and alternative baselines). Our model is capable of translating the human badly drawn sketch with detecting the desired facial structures (See Figure \ref{fig:badsketchtoimage}), and then perform  attribute editing and intensity control.
 
\end{itemize}

\begin{figure*}[!t]
    \centering
    \vspace{-2em}
    \includegraphics[width = 0.95\linewidth]{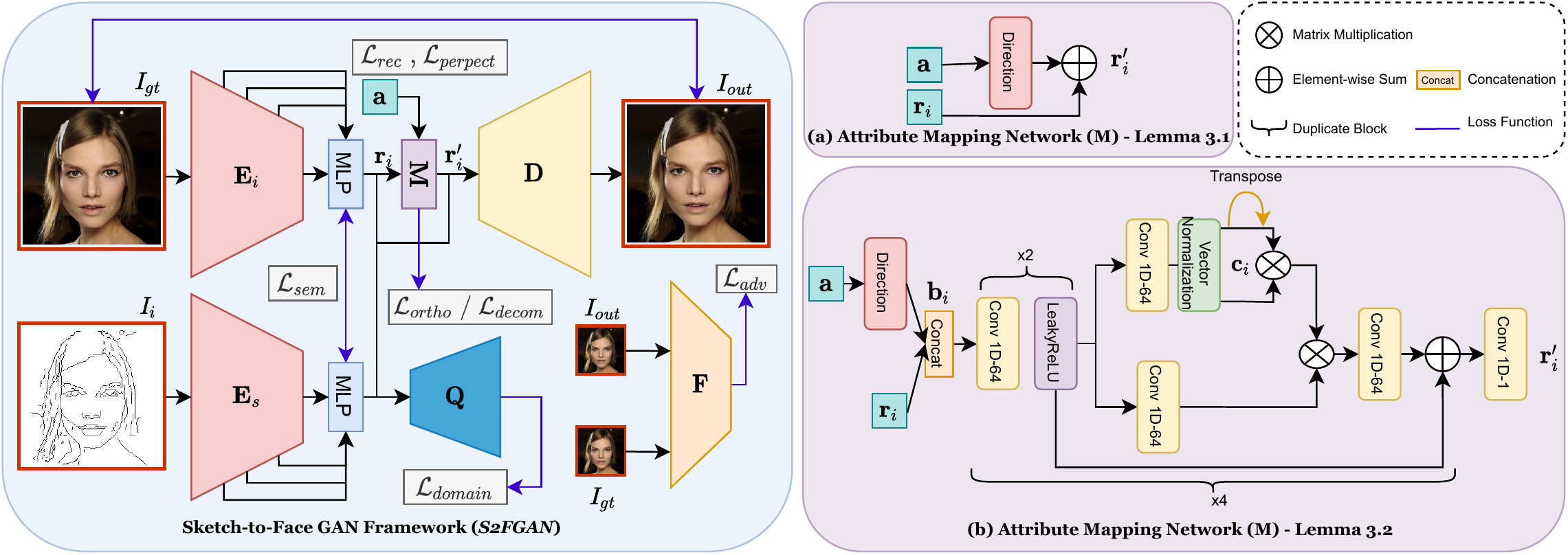}
    \vspace{-1em}
    \caption{\small The network architecture of \textit{S2FGAN} framework. The ground-truth face, $\mathit{I_{gt}}$ is projected to the latent code via Image Encoder, $\mathbf{E}_{i}$. The Sketch Encoder, $\mathbf{E}_{s}$ learns to map latent code, $\mathbf{E}_{i}(\mathit{I_{gt}})$ via minimizing the semantic perceptual loss. A domain discriminator, $\mathbf{Q}$ encourages $\mathbf{E}_{s}(\mathit{I_{i}})$ to be indistinguishable from ground-truth image latent code. Then, Attribute Mapping Networks enables interactive semantic editing from $\mathbf{a}$. The discriminator structure provides necessary supervision to support the generator. There are two versions of our Attribute Mapping Network.
    \textbf{(a)} After disentangling attributes semantics by using $\mathcal{L}_{ortho}$, we build the attribute editing by using Lemma \ref{lemma:ld}.
    \textbf{(b)} By modeling Lemma \ref{lemma:lc} with $\mathcal{L}_{decom}$, we can decompose latent code semantic with a goal of attribute editing.
    }
    \label{fig:netarch}
\end{figure*}

\section{Related Work}
\noindent\textbf{Sketch-to-Image Generation.} Generative Adversarial Networks (GANs) have shown great potential in computer vision tasks such as high-resolution image synthesis \cite{PGGAN,StyleGAN,StyleGAN2,pixel2pixelHD}, image completion \cite{SCGAN,DPS,FS}, image translation \cite{ContextGAN,DPS,FS,InterFaceGAN,P2P,StarGAN,AttGAN,FaderNetwork,unit,munit,latentSpaceInterplotation,GLO,rgeomotry,DFI,ALD}, and conditional image synthesis \cite{cGAN,ContextGAN,SCGAN,FS}. Among these applications, face-to-face translation is one of the most widely studied tasks, because faces carry influential social cues essential for human communication. This enables non-artists to simulate diverse images by sketching their abstract ideas. The goal is to map the sketch to their corresponding ground truth images \cite{DPS,ContextGAN,pixel2pixelHD,P2P,s2i}. Previous work such as Deep Plastic Surgery \cite{DPS} and ContextualGAN \cite{ContextGAN} had a target of increasing the model robustness by adapting poorly drawn sketches. However, they did not consider improving the generation process by including semantic descriptions from users. As a result, they could not control the diversity of image attributes. This paper aims to help the sketch-to-image generation processes by providing the opportunities to control intensities of attributes.

\noindent\textbf{Attribute Editing.} In Face-to-Face translation, attribute editing attempts to change certain features (e.g., big lips/big smile) of a given face, preserving identity information. Some methods edit attributes of photo-realistic faces by using pencil sketch as input \cite{SCGAN,DPS}. Others edit the face attributes by keyword descriptions (e.g., Male and Young) \cite{StarGAN,AttGAN,FaderNetwork,STGAN}. Because it requires less manual interaction, this paper explores the latter approach. A known issue about keyword-based editing is that it requires manually specifying both attributes to be edited and preserved \cite{AttGAN,StarGAN,FaderNetwork}. Recently, STGAN \cite{STGAN} addressed this issue by improving the generation process of AttGAN \cite{AttGAN}. However, they did not consider assisting the generation process of diverse image-to-image translation problems by involving semantic attributes. In another work, InterfaceGAN \cite{InterFaceGAN} successfully included a control on diverse image generation by editing the latent code proposed in StyleGAN \cite{StyleGAN} or PGGAN \cite{PGGAN}. However, they have weak GAN inversion results and lack analysis and effect on preserving multiple attributes that do not want to change.  All existing works of attribute editing operate on the domain of photo-realistic images, which lacks generalization ability. This paper focuses on attribute editing in the absence of photo-realistic images (e.g., sketch or edges) that can help users express their abstract ideas by compensating for the limited representation power of sketches, masks, and so on.

\section{Method}
\noindent \textbf{Problem Formulation.} Suppose, $\mathbf{A}$ represents the set of valid attributes (i.e. smile, old and so on) of faces. Given a sketch image $\mathit{{I}_{i}} \in \mathbb{R}^{H \times W}$ and desired attributes shifting vector $\mathbf{a} = [a_1, a_2 \ldots a_{|\mathbf{A}|}] \in \mathbb{R}^{|\mathbf{A}|}$, our goal is to find a parameterized generator $\mathbf{G}$ that learns a mapping, $\mathbf{G}(\mathit{{I}_{i}},\mathbf{a}) \rightarrow \mathit{{I}_{out}}$ from the sketch, $\mathit{{I}_{i}}$ to photo-realistic image, $\mathit{{I}_{out}}\in \mathbb{R}^{H \times W \times 3}$ described by the attributes shifting vector $\mathbf{a}$. By manipulating $\mathbf{a}$, users can control intensity of attributes in a generated face $\mathit{{I}_{out}}$. When $\mathbf{a = 0} \in \mathbb{R}^{|\mathbf{A}|}$, $\mathbf{G}(\mathit{I_{i}}, \mathbf{0})$ reconstructs the ground truth photo-realistic image $\mathit{{I}_{gt}}$. Otherwise, $\mathbf{G}(\mathit{{I}_{i}},\mathbf{a})$ generates manipulated photo-realistic face according to the value of $\mathbf{a}$. Shifting the value $a_{j}$ manipulates the $j$th attribute of face. Users have the flexibility to manipulate single or multiple attributes $a_j$ simultaneously where $j = 1,2\ldots |\mathbf{A}|$.

\begin{figure*}[!t]
    \centering
    \vspace{-1em}
    \includegraphics[width = 0.9\linewidth]{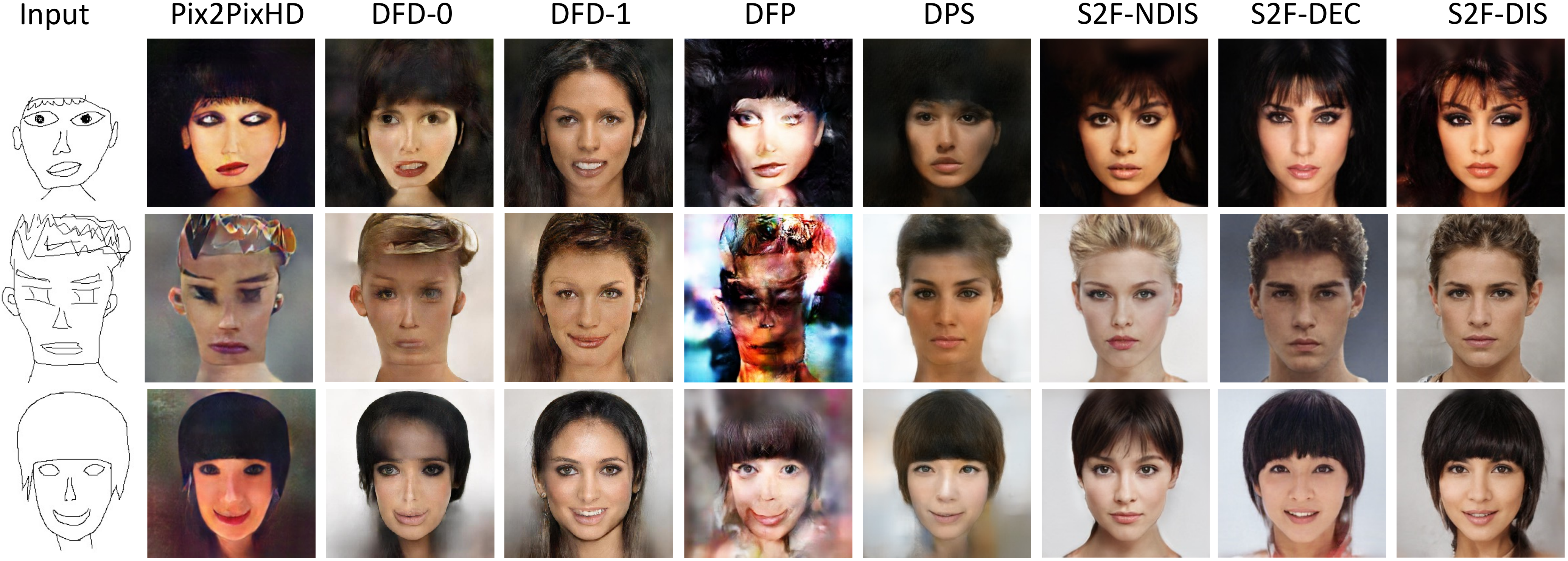}
    \vspace{-1em}
    \caption{\small Comparison of translating human-drawn sketches \cite{DPS} with Pix2PixHD \cite{pixel2pixelHD}, DeepFaceDrawing (DFD) \cite{DeepFaceDrawing}, DeepFacePencil (DFP) \cite{deepfacepencil} and Deep Plastic Surgery (DPS) \cite{DPS}. DFD-0 and DFD-1 represent the medium refinement and fully refinement of DFD.}
    \label{fig:badsketchtoimage}
\end{figure*}

\subsection{Attribute Editing on Latent Code}
We further decompose the generator, $\mathbf{G}(\cdot)$ into encoder, $\mathbf{E}(\cdot)$, attribute manipulation network, $\mathbf{M}(\cdot)$ and decoder, $\mathbf{D}(\cdot)$. The relations are defined below:
{\small
\begin{align*}
    \mathbf{r}_{i}  = \mathbf{E}(\mathit{{I}_{i}}), \quad \mathbf{r}_{i}' = \mathbf{M}(\mathbf{r}_{i}, \mathbf{a}), \quad \mathit{{I}_{out}} = \mathbf{D}(\mathbf{r}_{i}')
\end{align*}
}%
Where $\mathbf{r}_{i}$ and $\mathbf{r}_{i}'$ $\in \mathbb{R}^{d}$ and $d$ is the dimensionality of the latent code. We assume, $S^{+} \cup S^{-} = \mathbf{A}$, where, $S^{+}$ contains the set of attributes to be edited, and  $S^{-}$ are the rest of attributes in $\mathbf{A}$.

\begin{definition}
$[\mathbf{w}_{1} \ldots \mathbf{w}_{|\mathbf{A}|}]$ defines a list of hyperplanes that classifies $|\mathbf{A}|$ attributes. Each of the hyperplane is a unit vector such that $\forall_{i \in [1 \ldots |\mathbf{A}|]} \lVert \mathbf{w}_{i} \rVert^2 = 1$. The scalar product of $\mathbf{r}_{i}$ and $ \mathbf{w}_{i}$ defines the intensity of $i_{th}$ attribute. $\mathbf{w}_{i}^{T}\mathbf{r}_{i} \geq 0$ means that $i_{th}$ attribute is present in input image and otherwise absent.
\end{definition}

\noindent\textbf{Remark.} Face identify information is a summary of different facial attributes. Attribute editing with preserving the identity could be a paradox. For example, shifting attributes of a ``female'' to a ``male'' face loses the feminine identity of the input. Thus, we assume that the attribute editing operation preserves identity information if and only if the operation is invertible and effective when a sufficiently large dataset is available. Mathematically,
{\small
\begin{align*}
   \mathbf{r_{i}} = \mathbf{M}(\mathbf{M}(\mathbf{r}_{i}, \mathbf{a}), -\mathbf{a}), \,\,\, \mathbf{r}_{i} \mathbf{w}_{j} + \mathbf{a} =  \mathbf{M}(\mathbf{r}_{i}, \mathbf{a}) \mathbf{w}_{j}  \,\,\, 
    \forall j \in [1\ldots |\mathbf{A}|]
\end{align*}
}%
We propose two different ways to accomplish attribute editing. Firstly, we learn a disentangled latent space that forces all the attributes in set $\mathbf{A}$ to be orthogonal with each other. Editing a subset of attributes in $\mathbf{A}$ by simple additions
would not affect the rest of the attributes. Lemma \ref{lemma:ld} confirms the claim.
\begin{lemma}\vspace{-.25em}
Given the attributes shifting vector $\mathbf{a} = [a_1, a_2 \ldots a_{|\mathbf{A}|}]$. If $\forall j,k \in |\mathbf{A}| \times |\mathbf{A}|$ S.T. $\mathbf{w}_{i}$ is orthogonal with $\mathbf{w}_{j}$ and all hyperplanes correctly classify the input latent code $\mathbf{r}_{i}$, then we can have a disentangled latent space. Specifically, we can editing the attributes without affecting the rest attributes by $\mathbf{r}_{i} + a_{j}\mathbf{w}_{j}$.
\begin{proof}
We proof this claim by contradiction. Assume, editing the set of attributes $S^{+}$ will affect the set of attributes $S^{-}$. Then,
{\small
\begin{align}
    (\mathbf{r_{i}} + \sum_{j \in S^{+}} a_{j} \mathbf{w}_{j}) \mathbf{w}_{k} &\neq \mathbf{r_{i}} \mathbf{w}_{k}    \quad \forall k \in S^{-}\\
    \mathbf{r_{i}}\mathbf{w}_{k} + \sum_{j \in S^{+}} a_{j} \mathbf{w}_{j} \mathbf{w}_{k} &\neq \mathbf{r_{i}} \mathbf{w}_{k} \quad \forall k \in S^{-} \label{eq:p1}\\
    \mathbf{r_{i}} \mathbf{w}_{k} &\neq \mathbf{r_{i}} \mathbf{w}_{k} \quad \forall k \in S^{-} \label{eq:p2}
\end{align}
}%
Eq. \ref{eq:p2} can be inferred from Eq. \ref{eq:p1} because all hyperplanes are orthogonal with each other. Eq. \ref{eq:p2} derives the contradiction.
\vspace{-1.2em}
\end{proof}
\label{lemma:ld}
\end{lemma}

\noindent Secondly, attribute editing can be accomplished via semantic decomposition. We can edit the attributes in set $S^{+}$ by preserving the intensity of attributes in set $S^{-}$. However, the semantic of attributes in $S^{-}$ may change.  Lemma \ref{lemma:lc} describes detailed proves.

\begin{figure*}[!t]
    \centering
    \vspace{-1em}
    \includegraphics[width = 0.95\linewidth]{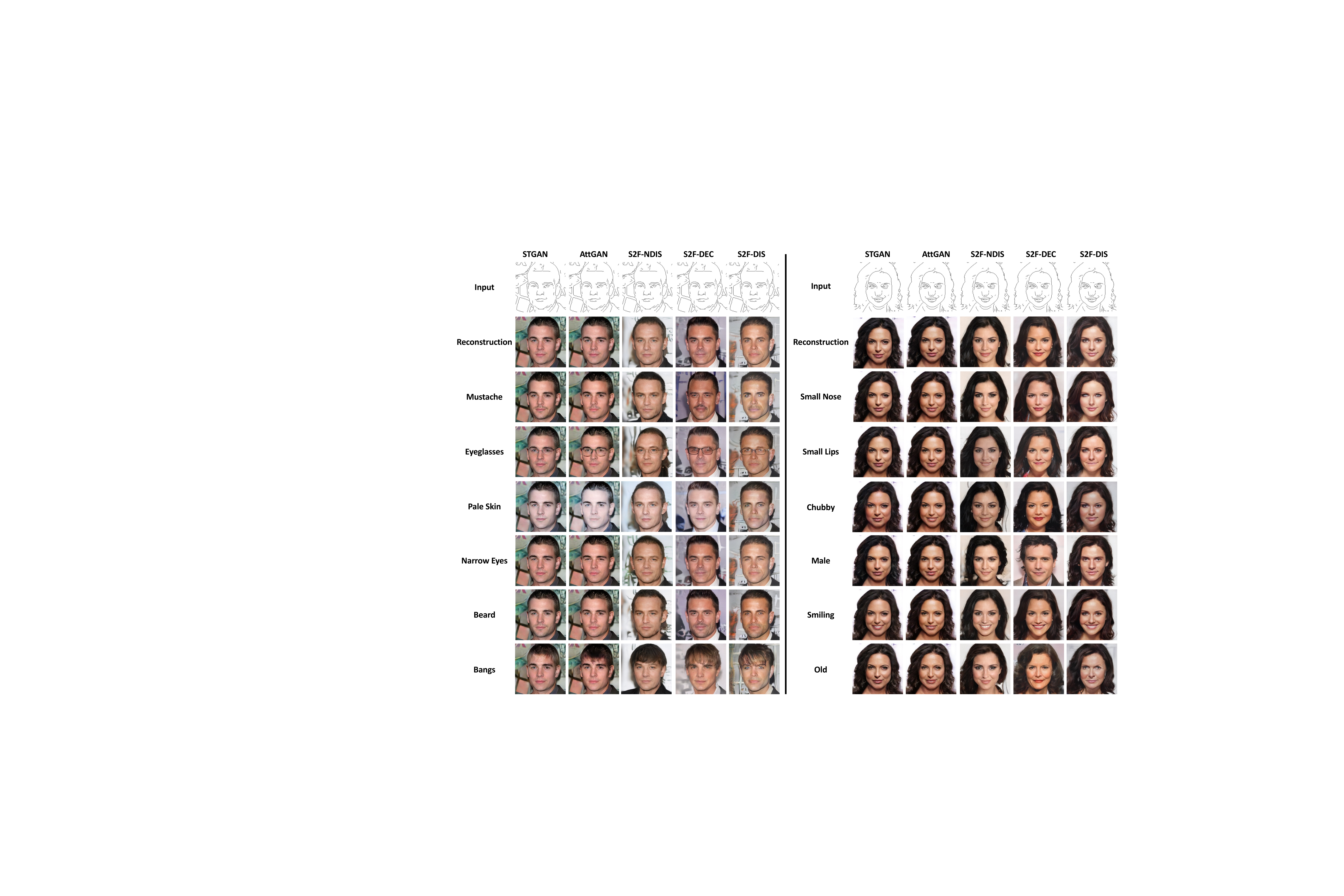}
    \vspace{-1em}
    \caption{\small Single attribute editing results for STGAN \cite{STGAN}, AttGAN \cite{AttGAN},  S2F-NDIS, S2F-DEC and S2F-DIS.}
    \label{fig:singleattribute}
\end{figure*}

\begin{lemma}
Given the attributes shifting vector $\mathbf{a} = [a_1, a_2 \ldots a_{|\mathbf{A}|}]$. There exists a linear  decomposition $\Gamma = [\Gamma_{0}(\cdot), \Gamma_{1}(\cdot) \ldots]$ S.T.  $ \mathbf{r}_{i} = \sum_{k \in |\Gamma|} \Gamma_{k}(\mathbf{r}_{i})$ and $\forall j \in |\mathbf{A}|, \mathbf{w}_{k} = \sum_{k \in |\Gamma|} \Gamma_{k}(\mathbf{w}_{j})$, where $\Gamma_{k}(\cdot)$ is a index function.
Denoting $\Gamma^+$ be the set of component need to be varied when editing attributes $S^{+}$. Then we can formulate the attribute editing below,
{\small
\vspace{-.5em}\begin{align}
    \Big(\sum_{j \in S^+} \sum_{k \in |\Gamma^+|} \eta_{k} \Gamma_{k}(\mathbf{w}_{j}) \Big) \mathbf{w}_{l} & = 0 &\quad \forall l \in S^{-} \label{eq:f1}\\
    \Big(\sum_{j \in S^+} \sum_{k \in |\Gamma^+|} \eta_{k} \Gamma_{k}(\mathbf{w}_{j})\Big) \mathbf{w}_{l} & = a_{l} &\quad \forall l \in S^{+} \label{eq:f2}
\end{align}
}%
$\eta_{k}$ is a scalar variable that varies the magnitude of sub-semantic $\Gamma_{k}(\mathbf{w}_{j})$. Eq. \ref{eq:f1} constrains that intensity scores of $S^-$ will be preserved during attribute editing. Eq. \ref{eq:f2} constrains that the intensity score of $S^+$ should be changed with expectation. The Eq. \ref{eq:f1} and Eq. \ref{eq:f2} have feasible solutions if and only if $d >= |\mathbf{A}|$. It means we can decompose the attribute semantic into different sub-semantics. Then, we can manipulate the attributes of interests $S^{+}$ by preserving the intensity of other attributes $S^{-}$ and modifying the magnitude of sub-semantics. 
\label{lemma:lc}
\end{lemma}

\subsection{\textit{\textbf{S2FGAN}} Framework}
We introduce our proposed framework, \textit{S2FGAN}, in Figure \ref{fig:netarch}. It has four components: a Encoder (Image Latent Encoder, $\mathbf{E}_{i}$, Sketch Latent Encoder, $\mathbf{E}_{s}$), Attribute Mapping Network, $\mathbf{M}$, Style Aware Decoder, $\mathbf{D}$ and a Discriminator, $\mathbf{F}$. For clarification, we use $\mathbf{G}_{i}$ and $\mathbf{G}_{s}$ to denote the image reconstruction branch $\mathbf{D}(\mathbf{M}(\mathbf{E}_{i}(I_{gt}), \mathbf{a}))$ and sketch-to-image generation branch $\mathbf{D}(\mathbf{M}(\mathbf{E}_{s}(I_{i}), \mathbf{a}))$ of our generation. Our sketch-to-image translation with attribute editing relies on encoding and manipulating the sketch's latent code. Our decoder and discriminator adopt StyleGAN \cite{StyleGAN2} as backbone. Here, we describe the architecture of the encoder and attribute mapping network.

\noindent \textbf{Encoder.} Regardless of any input sketch,  $\mathit{I}_{i}$, learning a direct mapping to photo-realistic image, $\mathit{I}_{out}$ is difficult. With the increase of the input's spatial resolution, the network should learn more fine-grained facial features to construct a photo-realistic look. However, the sketches often fail to describe such facial texture. To solve this problem, we design an Image Latent Encoder, $\mathbf{E}_{i}$ and a Sketch Latent Encoder, $\mathbf{E}_{s}$, that treat image construction as an auxiliary task to help the sketch-to-image generation. Our encoder is a simple $(\floor{\sqrt{\log(HW)}} - 2)$ layers ResNet \cite{resnet}. Considering different facial attribute desires different sizes of convolution feature maps, we pool the feature maps and pass the summarised feature to a multilayer perceptron (MLP) before each down-sampling operations. We then sum the processed feature from different encoder hierarchies and refine these features through another MLP. These operations enable the encoder to learn the facial feature spatially. For example, a pale skin attribute would more reasonably present in early convolution features of the encoder. \\
\noindent \textbf{Attribute Mapping Network.} 
We describe two different versions of Attribute Mapping Network using Lemma \ref{lemma:ld} and Lemma \ref{lemma:lc}. The objectives for training are presented in the next subsection. 

\noindent\textit{According to Lemma \ref{lemma:ld}}, after disentangling the attribute semantics of interest by ensuring them to be orthogonal to each other, the attribute editing task becomes a simple addition operation. We perform the attribute editing by Eq. \ref{eq:ed} in Figure \ref{fig:netarch} (a).
{\small
\begin{align}
    \mathbf{r}_{i}' = \mathbf{r}_{i} + \sum_{j \in [1 \ldots |\mathbf{A}|]} a_{j} \mathbf{w}_{j}
    \label{eq:ed}
\end{align}
}%
\noindent\textit{According to Lemma \ref{lemma:lc}}, we model Eq. \ref{eq:f1} and Eq. \ref{eq:f2} inside the subnetwork Attribute Mapping Network (Figure \ref{fig:netarch} (b)) which maintains a trainable embedding layer $\mathbf{e}$, where $\mathbf{e} \in \mathbb{R}^{|\mathbf{A}| \times d}$. This embedding layer (named Direction in Figure \ref{fig:netarch}) could also be a replica of attributes hyperplane $[\mathbf{w}_{1} \ldots \mathbf{w}_{|\mathbf{A}|}]$. Let $f_{1}$, $f_{2}$, $f_{3}$ and  $f_{4}$ denote four convolution 1D layer. And $\delta$ denote a multi-layer convolution block, which models semantic decomposition. The key operations of attribute mapping network are:
{\small
\begin{align*}
    \mathbf{b}_{i}  &= \delta(concat(\mathbf{e} \cdot \mathbf{a}, \mathbf{r}_{i})), \quad \mathbf{c}_{i}  = \mathbf{Norm}(f_{1}(\mathbf{b}_{i})),\\
    \mathbf{r}_{i}' &=  f_{2} (\mathbf{c}_{i} \mathbf{c}_{i}^{T} f_{3}(\mathbf{b}_{i})) + f_{4}(concat(\mathbf{e} \cdot \mathbf{a}, \mathbf{r}_{i}))
\end{align*}
}%
We repeat operations above four times. $concat$ and $\mathbf{Norm}$ represent channel-wise concatenation and vector normalization along channel dimension, respectively.  $\mathbf{b_{i}}$ and $\mathbf{c_{i}}$ are intermediate variables. First, $\delta$ learns to decompose the current latent code and desired attribute shifting latent code into sub-semantics. Second, we calculate the cosine similarity between each of the sub-semantics, which scores each unit semantic. It aims to incorporate the magnitude changes for different sub-semantics for preserving the intensity of non-modified attributes. For attributes with non-zero attribute shifting value, they may correlate with each other, and the sub-semantics among them also need to be aligned. Meanwhile, the best-edited representation is searched with the awareness of the current latent code $\mathbf{r}_{i}$. This idea (Lemma \ref{lemma:lc}) is especially helpful when multiple attribute editing is involved in a latent space, and a pretrained generator (encoder and decoder) is available. In this way, we can adapt to arbitrary complex attribute editing while preserving the appearance of the sketch $\mathit{{I}_{i}}$.

\begin{figure*}[!t]
    \centering
    \vspace{-1em}
    \includegraphics[width = 0.90 \linewidth]{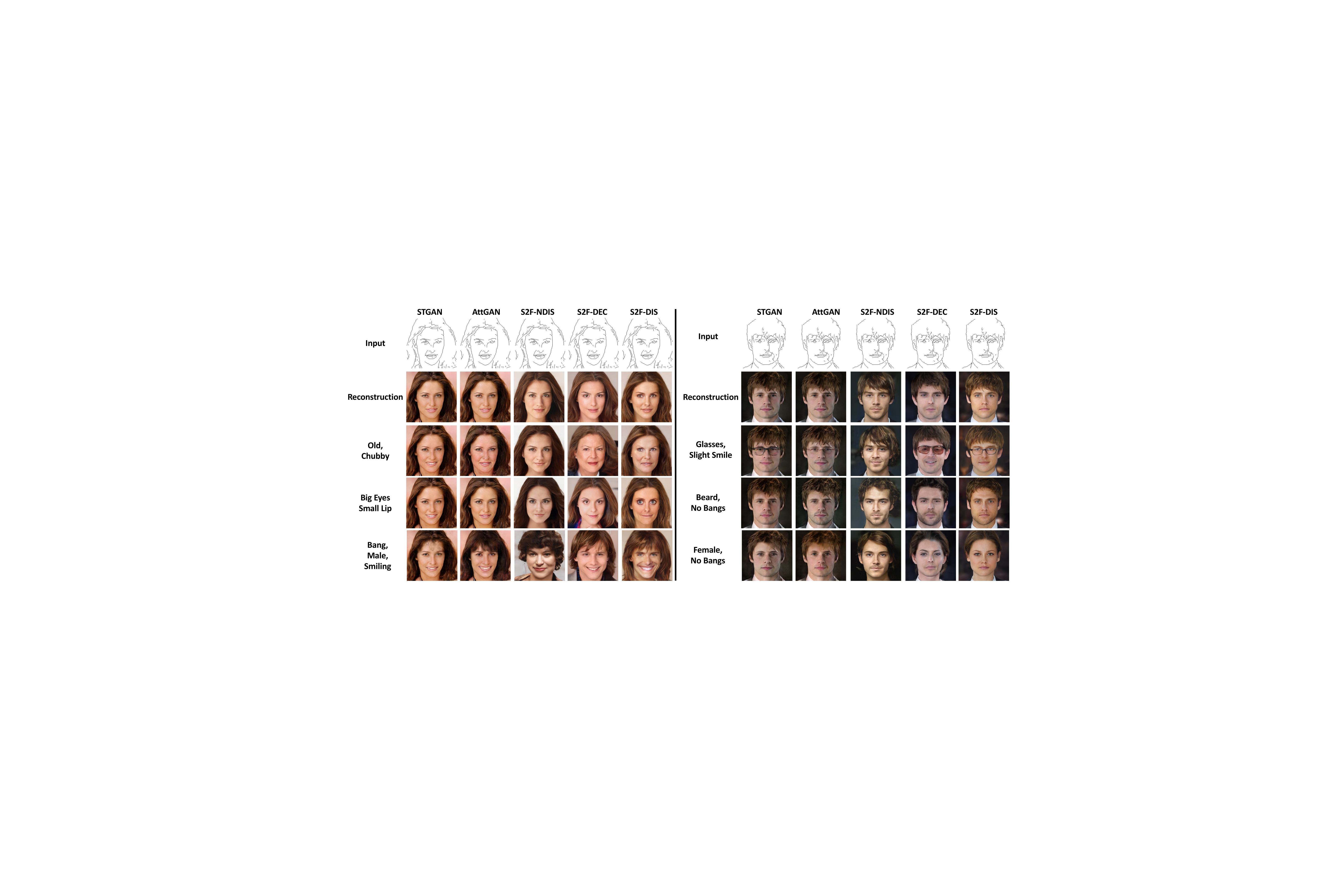}
    \vspace{-1em}
    \caption{\small Multi-attribute editing results for AttGAN \cite{AttGAN}, STGAN \cite{STGAN}, S2F-NDIS, S2F-DEC and S2F-DIS.}
    \label{fig:multiattributecompare}
\end{figure*}

\subsection{Multi-Objective Learning}
There are two extra objective functions in our compared with conventional GAN framework \cite{StyleGAN,StyleGAN2}. Let $\mathbf{u}_{i}$ denote attributes of ground truth images $\mathit{{I}_{gt}}$.

\noindent \textbf{Semantic Level Perceptual Loss.} If the attribute shifting vector, $\mathbf{a = 0} \in \mathbb{R}^{|\mathbf{A}|}$, the generator, $\mathbf{G}_{s}$ will reconstruct the ground-truth image, $\mathit{{I}_{gt}}$ from a sketch input, $\mathit{I_{i}}$. Instead of directly imposing the L1 loss \cite{P2P,munit,AttGAN,StarGAN,STGAN}, Perceptual Loss \cite{pixel2pixelHD, DPS}, and Feature Matching Loss \cite{pixel2pixelHD,DPS,DeepFaceDrawing,deepfacepencil} between the synthesis $\mathit{I_{out}}$ and ground-truth $\mathit{I_{gt}}$ image, we propose to match the semantics of synthesis and ground-truth image on the latent space. With the increase of desired synthesis resolution, the size of model output increases exponentially. In that case, optimization using L1, perceptual, and feature matching loss become more challenging. In contrast, our GAN's latent space calculates a discriminative summarization of low-dimensional image features. It maintains domain-specific perceptual context and ensures faster optimization. We define the Semantic Level Perceptual Loss as,
{\small
\begin{align}
    \mathcal{L}_{sem} = \mathbb{E} \big[ \lVert \mathbf{E}^{*}_{i}(\mathit{I_{gt}}) - \mathbf{E}_{s}(\mathit{I_{i}})  \rVert_{2} \big]
\end{align}
}%
where '*' indicates the component removed from the computation graph during backpropagation. A domain discriminator $\mathbf{Q}$ regularizes the sketch encoder $E_{s}$. It encourages sketch latent code, $\mathbf{E}_{s}(\mathit{I_{i}})$ to stick around the latent space of $E_{i}(\mathit{I_{gt}})$. This is helpful when translating the bad drawn sketches. (See Figure \ref{fig:badsketchtoimage}). To avoid two-stage adversarial training, GRL layer \cite{dnn} is used to reversal the gradient before updating the sketch encoder $\mathbf{E}_{s}$ during training.
{\small
\begin{align}
    \mathcal{L}_{domain} &= \log \Big(1 - \mathbf{Q}\big(\mathbf{E}_{s}(\mathit{I_{i}}))\big) +  \log(1 - \mathbf{Q}\big(\mathbf{M}(\mathbf{E}_{s}(\mathit{I_{i}}), \mathbf{a})\big) \Big) \quad \nonumber\\
    & + \log \mathbf{Q}\big( \mathbf{E}^{*}_{i}(\mathit{I_{gt}})\big) +  \log \mathbf{Q} \big(\mathbf{M}(\mathbf{E}^{*}_{i}(\mathit{I_{gt}})), \mathbf{a}\big) 
\end{align}
}%
Meanwhile, to discover the latent code of ground-truth image $\mathit{{I}_{gt}}$, we use image reconstruction as an auxiliary task. Another intuition is that image reconstruction is simpler than sketch-to-image translation. We use the L1-Loss $\mathcal{L}_{rec}$ and perceptual loss $\mathcal{L}_{percept}$ for image reconstruction.

\begin{figure*}[!t]
    \centering
    \vspace{-1em}
    \includegraphics[width=0.85\linewidth]{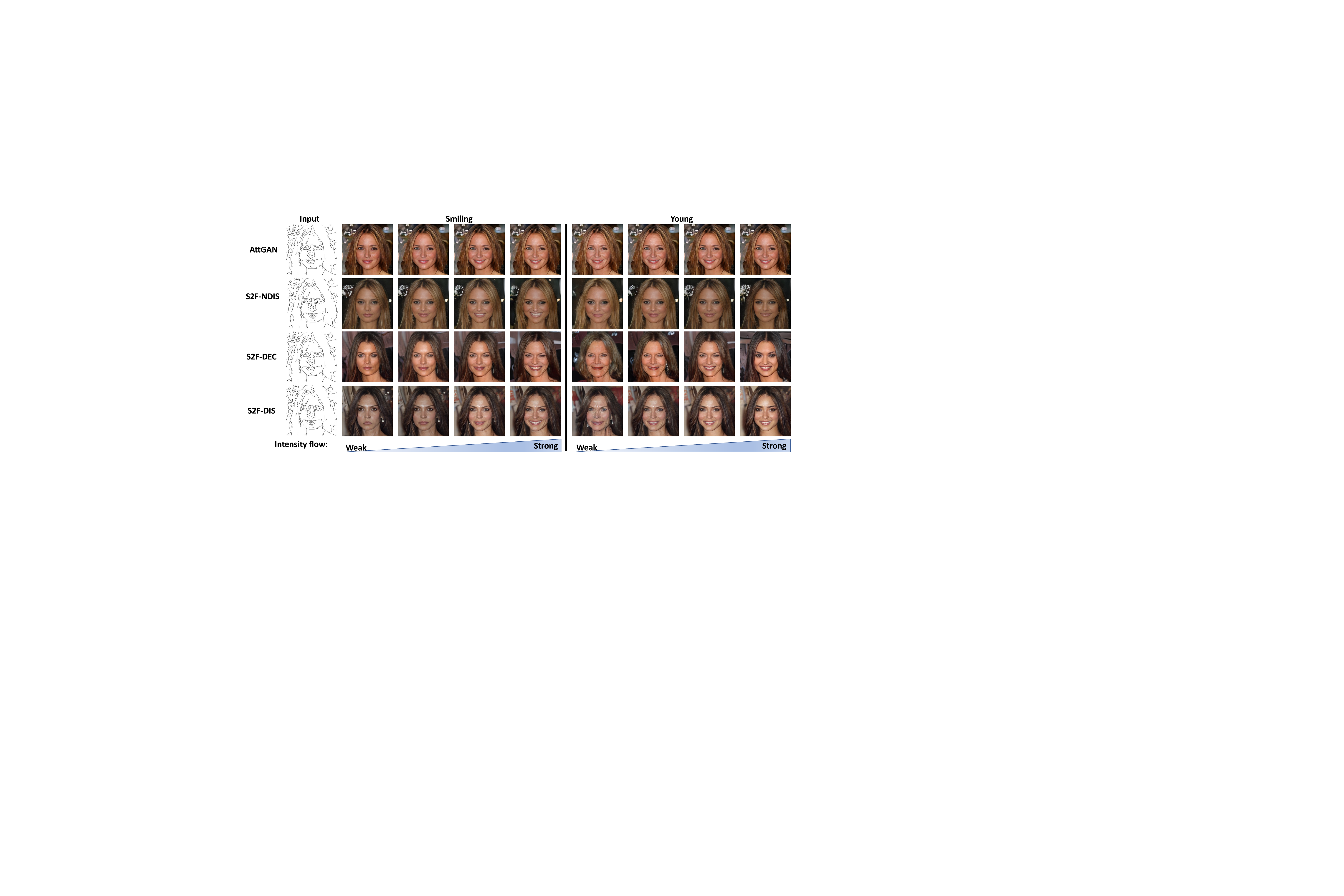}
    \vspace{-1em}
    \caption{\small Intensity control for AttGAN \cite{AttGAN},  S2F-NDIS, S2F-DEC and S2F-DIS on two state-like attributes (Smiling and Young).}
    \label{fig:intensitycontrol}
\end{figure*}
\noindent \textbf{Latent Semantic Loss.} As classifying the attributes from the sketch is noisy, we leverage the side effect of auxiliary image reconstruction tasks. Here, we describe the attribute reconstruction loss corresponding to Lemma \ref{lemma:ld} and Lemma \ref{lemma:lc}, separately.\\
For Lemma \ref{lemma:ld}, we focus on learning a disentangled latent space, where all attributes in $\mathbf{A}$ are orthogonal to each other. The loss is:
{\small
\begin{align}
    \mathcal{L}_{ortho} = \mathbb{E} \big[\lVert \mathbf{E}_{i}(\mathit{I_{gt}}) \mathbf{W}^{T} - \mathbf{u}_{i} \rVert^{2}_{2} \big] + \lVert \mathbf{W}\mathbf{W}^{T} - \mathbbm{1}\rVert_{2}
\end{align}
}%
We denote $\mathbf{W} = [\mathbf{w}_{1} \ldots \mathbf{w}_{|\mathbf{A}|}] \in \mathbb{R}^{|\mathbf{A}| \times d}$ as trainable parameters and $\mathbbm{1}$ as a $|\mathbf{A}| \times |\mathbf{A}|$ identity matrix.\\
\noindent\textbf{Remark.} We have to maintain a proper dimensionality $d$ for the latent code to prevent the co-adaption of $\mathcal{L}_{rec}$ and $\mathcal{L}_{ortho}$. If $d$ is extremely large, they ($\mathcal{L}_{rec}$ and $\mathcal{L}_{ortho}$) will depend on different latent code chunks, leading to the failure of further editing of attributes. In our experiment, we set $d$ to 512 during our training.\\
For Lemma \ref{lemma:lc}, we aim to learn a decomposition that preserves attributes' intensity when it is not edited. The associated loss is:
{\small
\begin{align}
    \mathcal{L}_{decom} = & \hspace{1em} \mathbb{E} \big[\lVert \mathbf{E}^{*}_{i} (\mathit{I_{gt}}) \mathbf{W}^{T} - \mathbf{u}_{i} \rVert^{2}_{2} \big] \quad  \nonumber\\
    & + \mathbb{E}\big[\lVert \mathbf{M}(\mathbf{E}^{*}_{i}(\mathit{I_{gt}}), \mathbf{a}) - \mathbf{E}^{*}_{i}(\mathit{I_{gt}}) (\mathbf{W}^{*})^{T} - \mathbf{a} \rVert^{2}_{2} \big] \quad \nonumber \\
    & + \mathbb{E}\big[\lVert \mathbf{M}(\mathbf{M}(\mathbf{E}^{*}_{i}(\mathit{I_{gt}}), \mathbf{a}),- \mathbf{a}) - \mathbf{E}^{*}_{i}(\mathit{I_{gt}}) \rVert_{2}  \big]
\end{align}
}%
The first component learns to classify the attributes from the latent code. The second component enforces the attribute mapping networks manipulating the attributes with the desired intensity. The third component ensures the identity-invariant of attribute editing, i.e., the latent code should be reconstructed with two inverse editing behaviors. There is a significant difference between $\mathcal{L}_{ortho}$ and $\mathcal{L}_{decom}$. $\mathcal{L}_{ortho}$ forces the encoder to decorrelate the attributes $\mathcal{A}$ with gradient support. $\mathcal{L}_{decom}$ learns to classify the attributes from latent code while removing the latent code from the computation graph. During our experiments, the $\mathcal{L}_{ortho}$ is stronger at attribute editing when the image quality of $\mathcal{L}_{decom}$ is better. $\mathcal{L}_{ortho}$ decorrelates the attribute semantics that contradicts with data distribution. In contrast, $\mathcal{L}_{decom}$ allows correlating attributes. Additionally, we use the conventional adversarial loss $\mathcal{L}_{adv}$ with gradient penalty $\mathcal{L}_{R1}$ from \cite{StyleGAN, StyleGAN2} to train our model.

\noindent \textbf{Final Objectives.} The Lemma \ref{lemma:lc} and \ref{lemma:ld} lead two distinct models. They use the same semantic level perceptual loss, adversarial loss, and gradient penalty but corresponding with different latent semantic loss as indicated above. See \textbf{supplementary material} for details.

\section{Experiments}
\noindent \textbf{Dataset.} We use a high-quality face dataset CelebAMask-HQ \cite{celebAMASK-HQ}. For this work, we resize all images to resolution $256 \times 256 $. We choose every 20th image as the testing set ($1,500$ images) from the whole dataset, and the remaining images are in the training set ($28,500$ images). We experiment with twelve state-like attributes (Smiling, Male, No\_Beard, Eyeglasses, Young, Bangs, Narrow\_Eyes, Pale\_Skin, Big\_Lips, Big\_Nose, Mustache, and Chubby). The statistic of each attribute in the training and testing set are summarised in supplementary materials.To obtain face sketches, we use HED edge detector \cite{HED}, and post-process with the steps employed by Isola \textit{et al.} \cite{P2P}. We use synthesized real drawings from edge-based sketches \cite{DPS,P2P,SCGAN,FS} to avoid laborious costs and difficulty in collecting hand drawings and images pair.

\noindent \textbf{Method for Comparison.} \textit{For sketch-to-image translation}, we compare our model with Pix2PixHD \cite{pixel2pixelHD}, Deep plastic Surgery (DPS) \cite{DPS}, Deep Face Drawing (DFD) \cite{DeepFaceDrawing} and Deep Face Pencil (DFP). \textit{For attribute edining}, we compare our work with the most recent model-based state-of-the-art attribute generation methods, AttGAN , and STGAN. These methods are originally designed to manipulate attributes of photo-realistic images. For this, we train a Pix2PixHD model \cite{pixel2pixelHD} to translate sketches to images before applying attribute manipulation. The local enhancer of Pix2PixHD \cite{pixel2pixelHD} are removed to compare in resolution $256 \times 256$. Thus, we train all models from scratch with their published recommended settings if original authors do not provide checkpoints in the CelebAMask-HQ dataset.
We denote \textit{our work} as S2F-DIS and S2F-DEC motivated from Lemma \ref{lemma:ld} and Lemma \ref{lemma:lc}, respectively. DIS and DEC are short for disentangle and decompose, respectively. They are our core concepts for the two lemmas. We also compare the final recommendation of our approach with several plausible alternatives. To illustrate the effectiveness of the proposed image reconstruction auxiliary task and semantic perceptual loss, we remove the image encoder $\mathbf{E}_{i}$ of S2F-DIS while denoting this ablation with S2F-NDIS.\\ \noindent\textbf{Implementation Details.\footnote{Code and models are available at:  \url{https://github.com/Yan98/S2FGAN}}} We train our model from scratch under Adam optimizer with learning rate $0.002$. The decays of Adam optimizer are set to 0.0 and 0.99. Our model is trained on two Nvidia Tesla Volta V100-SXM2-32GB with Intel Xeon Cascade Lake Platinum 8268 (2.90GHz) CPUs. We use batch size 24. During training, images and corresponding sketches are randomly cropped and resized with ratio 0.8 and probability 0.5. We randomly used horizontal flip with probability of 0.5 also. When trains S2F-DIS, $\mathbf{a}$ always set to zero. See the \textbf{supplementary material} for detailed network architecture and parameters.

\newcommand*\rot{\rotatebox{90}}
\begin{table*}[!t]
 \begin{minipage}[!t]{0.68\linewidth}
    \centering
        \caption{\small Quantitative evaluation: facial attributes editing accuracy ($\%$). 
        }
        \vspace{-1em}
        \scalebox{.7}{\begin{tabular}{ccccccccccccc}
        \toprule
        \rot{Model} & \rot{Smiling} & \rot{Male} & \rot{\makecell{No\\Beard}} & \rot{\makecell{Eye\\glasses}} & \rot{Young} & \rot{Bangs}  & \rot{\makecell{Narrow\\Eyes}} & \rot{\makecell{Pale\\Skin}} & \rot{\makecell{Big\\Lips}} & \rot{\makecell{Big\\Nose}} & \rot{Mustache} & \rot{Chubby}  \\
        \midrule
        AttGAN \cite{AttGAN} &  56.0 & 24.0 & 44.0 & 82.0 & 46.0 & 56.0  & 80.0 & 66.0 & 66.0 & 54.0 & 10.0 & 12.0 \\ 
        STGAN \cite{STGAN} & 72.0 & 36.0 & 64.0 & 92.0 & 56.0 & 84.0  & 81.0 & 86.0 & 80.0 & 72.0 & 45.0 & 32.0 \\ 
        S2F-NIDS & \textbf{100} & \textbf{100} & 97.8 & 38.4 & 96.5 & 99.5 & \textbf{99.5} & 91.0 & 68.2 & 84.5 & \textbf{66.4} & \textbf{78.4} \\
        S2F-DEC & \textbf{100} & 99.6 & 57.3 & 99.3 & 98.8 & \textbf{99.8} & 99.4 & \textbf{96.2} & 81.3 & 92.1 &  63.9 & 75.5\\
        S2F-DIS & \textbf{100} & 99.9 & \textbf{98.9} & \textbf{99.7} & \textbf{99.5} & 98.0 & 91.4 & 72.5 & \textbf{90.0} & \textbf{98.9} & 66.1 & 58.3 \\
        \midrule
        Ground-truth & 93.0 & 97.6 & 95.8 & 99.5 & 87.9 & 94.0  & 91.0 & 95.7 & 68.7 & 80.3 & 95.9 & 94.4 \\ 
        \bottomrule
    \end{tabular}}
    \label{Table:eva}
    \end{minipage}
    \begin{minipage}[!t]{0.29\textwidth}
    \centering
    \caption{\small Quantitative evaluation of synthesis quality and diversity.}
    \vspace{-1em}
    \scalebox{.75}{\begin{tabular}{|l|c|c|}
    \hline 
    \diagbox{Model}{Metric}  & IS & FID \\
    \hline
    DFP \cite{DeepFaceDrawing} & $2.61 \pm 0.02$ & $339.2 \pm 0.85 $  \\
    \hline
    DFD \cite{deepfacepencil} & $\mathbf{3.09} \pm 0.03$ & $313.8 \pm 0.53$ \\
    \hline
    DPS \cite{DPS} & $2.65 \pm 0.03$ & $325.7 \pm 0.88$ \\
    \hline
    Pix2PixHD \cite{pixel2pixelHD} & $2.99 \pm 0.04$  & $38.71 \pm 0.69$ \\
    \hline
    S2F-NDIS & $2.58 \pm 0.04$ & $34.18 \pm 0.37$\\
    \hline
    S2F-DEC & $2.73 \pm 0.03 $ & $28.21 \pm 0.76$ \\
    \hline
    S2F-DIS & $3.04 \pm 0.04$ & $\mathbf{26.14 \pm 0.43}$\\
    \hline
    \end{tabular}} 
    \label{Table:evascore}
    \end{minipage}
    \vspace{-1em}
\end{table*}

\subsection{Sketch-to-Image Translation}
\textbf{Translating Human Drawn Sketches.} We compare with the state-of-the-art models (with their provided checkpoint) by using the human drawn sketches \cite{DPS} in Figure \ref{fig:badsketchtoimage}. Although \cite{DPS,DeepFaceDrawing,deepfacepencil,pixel2pixelHD} used different sketch extraction methods, the ultimate goal is to adapt the human-drawn sketches to retrieve photo-realistic images. The Pix2PixHD, DFD, and DFP fail in generating high-quality and sensitive images. Though DPS and S2F-NDIS can capture the outline of input sketches, the image quality remains questionable. The layouts and shape variations between input sketch and output synthesis are expected because we subjectively defined the refinement level. Depends on users' confidence in their drawn sketches, they have the flexibility of controlling the fitness between input sketches and output synthesises (See Figure 4-7 in the \textbf{supplementary material} for synthesis without any refinement).
Our method combines the truncation trick \cite{truncationtrick} (used in StyleGAN \cite{StyleGAN,StyleGAN2}) and k-nearest neighborhood for adapting badly drawn sketches (used in DFP \cite{DeepFaceDrawing}). Because of the superior semantic level perceptual loss (with domain regularization), our sketch encoder $\mathbf{E}_{s}$ searches the sketches' latent code within the scope of ground truth image's $\mathit{I_{gt}}$ latent space. It helps the model to refine the badly drawn human sketches. See \textbf{supplementary material} for algorithm details, more comparison and style transfer in refining drawn sketches and comparison for translating machine extracted sketches.\\
\noindent\textbf{Attribute Editing.} We present the comparison of S2F-DIS, S2F-DEC, and baseline methods for single and multiple attribute editing. For the single attribute case (see Figure \ref{fig:singleattribute}), AttGAN \cite{AttGAN} and STGAN \cite{STGAN} fail in the majority of attribute editing cases and have weak editing effect, for example, for the ``Male'' case. We also compare our proposed architecture with AttGAN \cite{AttGAN} and S2F-NDIS for attribute intensity control in Figure \ref{fig:intensitycontrol}.  Here, we present the easiest editing examples, smiling and young. Because of learning the orthogonal semantic vectors, the S2F-DIS leads to a better editing performance. Meanwhile, this learning process needs to de-correlate the data, making the image quality slightly worse than the S2F-DEC. For multi-attribute editing cases (see Figure \ref{fig:multiattributecompare}), \cite{STGAN,AttGAN} sometimes ignores part of the attributes that need to be changed, such as ``Old'', and ``Chubby'', ``Beard'' and ``No Bang'' editing cases. Moreover, those methods usually fail to retouch the existing attributes such as to change ``Lip Size''.  AttGAN \cite{AttGAN} and STGAN \cite{STGAN} control the attribute editing based on concatenation of down-sampled inputs and scaled labels that lack continuity and variation. S2F-DEC aims to find a potentially attribute editing under Eq. \ref{eq:f1} and \ref{eq:f2}. It only constrains the intensity score of attributes, where attributes' semantic may change as we suggested in Lemma \ref{lemma:lc}. Thus, after attribute editing, there might be a different visual representation for all/some attributes but only guarantees the equivalence of intensity score between before- and after-editing. For S2F-DIS, semantic and intensity can be preserved simultaneously, as Lemma \ref{lemma:ld} indicates. However, both do not provide any performance guarantees for the attributes that are not in interest. S2F-NIDS lacks semantic level perceptual loss during training, making the learning noisy and leading to poor editing results. Our final recommendation (S2F-DEC, S2F-DIS) generates photo-realistic images from a sketch with superior attribute intensity control. We present more qualitative results in the \textbf{supplementary material.}
\subsection{Quantitative Evaluation.} 
In Table \ref{Table:eva}, we use a classifier suggested by He \textit{et.al} \cite{AttGAN} to validate the accuracy of attribute editing for our model. The evaluation classifier is trained on our training data of CelebAMask-HQ dataset \cite{celebAMASK-HQ} for resolution $128 \times 128$. We bilinearly downsample the generated images to resolution $128 \times 128$ before feeding them into the evaluation classifier. Note, there are around $\frac{1}{7}$ of training data where the target resolution is higher than their original implementation (compared with the original AttGAN \cite{AttGAN} and STGAN \cite{STGAN}). Our proposed framework S2F-DEC and S2F-DIS outperform the rest of the methods in the majority of cases. In the last row (GT) of Table \ref{Table:eva}, we also present the classifier's performance on real testing data that serves the upper bound of performance. Our model can control attribute intensity, which can construct obvious attributes for the evaluation. It also can enable the evaluation classifier to identify the edited attributes easily.
Moreover, we use Frechet Inception Distance (FID) \cite{FID-score} and Inception Score (IS) \cite{IS-score} to measure diversity and quality of synthesized images (1,500 testing images) in Table \ref{Table:evascore}. We do not set any refinement level for all methods during the comparison because it is a subjective metric. Similar to \cite{deepfacepencil}, we calculate the FID and IS based on the simulated human-drawn sketches by randomly dilating, deforming, and removing some edges from machine extracted sketches. DFD \cite{DeepFaceDrawing} and DPS \cite{DPS} heavily rely on defining appropriate refinement levels and hence not performing well.
Though S2F-DEC has worse IS than Pix2PixHD, S2F-DEC can better translate human-drawn sketches and provide advanced rendering facial textures (Figure \ref{fig:badsketchtoimage}). The S2F-DIS earns the best and second-best performance in FID and IS, respectively. To better understand the performance of our model, we present a user study in \textbf{supplementary material} for comparing with state-of-the-art methods.

\section{Conclusion}
This paper proposes two photo-realistic face generation models, S2F-DEC and S2F-DIS, given a sketch image as input. They can ascribe attributes on the generated face and include smooth manipulation over intensities of attributes. Further, our generated face preserves subject identity considering both single and multi-attribute editing cases. By adopting proposed semantic level perceptual loss and latent semantic loss, we can construct photo-realistic faces with the flexibility of shifting desired face attributes. Experiments on large face collection datasets demonstrate that S2F-DEC and S2F-DIS can accurately edit face attributes with more excellent controllability, even with non-photo-realistic inputs.

{\small
\bibliographystyle{ieee_fullname}
\bibliography{egbib}
}

\end{document}